%% file: main.tex
\documentclass[10pt,twocolumn,letterpaper]{article}

\usepackage{cvpr}              

\input{preamble}

\definecolor{cvprblue}{rgb}{0.21,0.49,0.74}
\usepackage[pagebackref,breaklinks,colorlinks,citecolor=cvprblue]{hyperref}

\usepackage{subfloat}
\usepackage{multirow}
\usepackage{colortbl}

\def\paperID{6135} 
\def\confName{CVPR}
\def\confYear{2024}

\begin{document}

\title{iKUN: Speak to Trackers without Retraining}

\author{Yunhao Du$^{1}$, Cheng Lei$^{1}$, Zhicheng Zhao$^{1,2,3}$\footnotemark[1]\ , Fei Su$^{1,2,3}$\\
$^1$The school of Artificial Intelligence, Beijing University of Posts and Telecommunications \\
$^2$Beijing Key Laboratory of Network System and Network Culture, China\\
$^3$Key Laboratory of Interactive Technology and Experience System Ministry \\of Culture and Tourism, Beijing, China\\
{\tt\small \{dyh\_bupt,mr.leicheng,zhaozc,sufei\}@bupt.edu.cn} \\
}
\maketitle
\renewcommand{\thefootnote}{\fnsymbol{footnote}}
\footnotetext[1]{Corresponding author}

\maketitle

\begin{abstract}
    Referring multi-object tracking (RMOT) aims to track multiple objects based on input textual descriptions.
    Previous works realize it by simply integrating an extra textual module into the multi-object tracker.
    However, they typically need to retrain the entire framework and have difficulties in optimization.
    In this work, we propose an \textbf{i}nsertable \textbf{K}nowledge \textbf{U}nification \textbf{N}etwork, termed \textbf{iKUN}, 
    to enable communication with off-the-shelf trackers in a plug-and-play manner.
    Concretely, a knowledge unification module (KUM) is designed to adaptively extract visual features based on textual guidance.
    Meanwhile, to improve the localization accuracy, 
    we present a neural version of Kalman filter (NKF) to dynamically adjust process noise and observation noise based on the current motion status.
    Moreover, to address the problem of open-set long-tail distribution of textual descriptions, 
    a test-time similarity calibration method is proposed to refine the confidence score with pseudo frequency.
    Extensive experiments on Refer-KITTI dataset verify the effectiveness of our framework.
    Finally, to speed up the development of RMOT, we also contribute a more challenging dataset, Refer-Dance, 
    by extending public DanceTrack dataset with motion and dressing descriptions.
    The codes and dataset are available at \url{https://github.com/dyhBUPT/iKUN}.
\end{abstract}

\begin{figure}
    \centering
    \subfloat[previous framework]{
        \includegraphics[width=.46\textwidth]{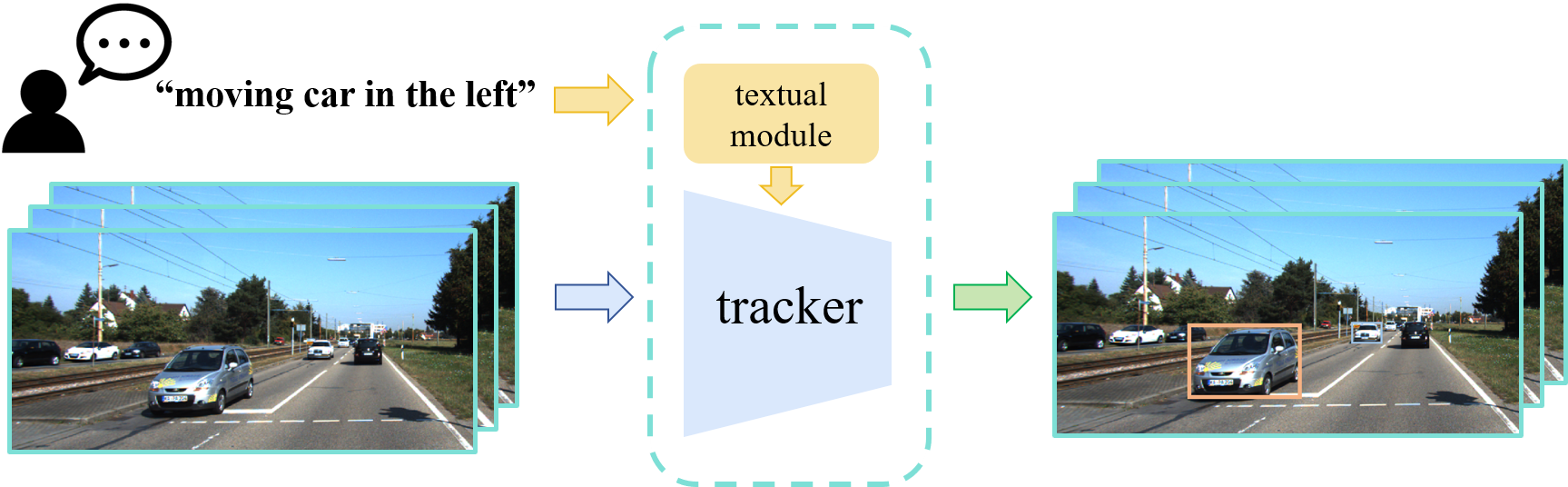}
    }\\
    \subfloat[our framework]{
        \includegraphics[width=.46\textwidth]{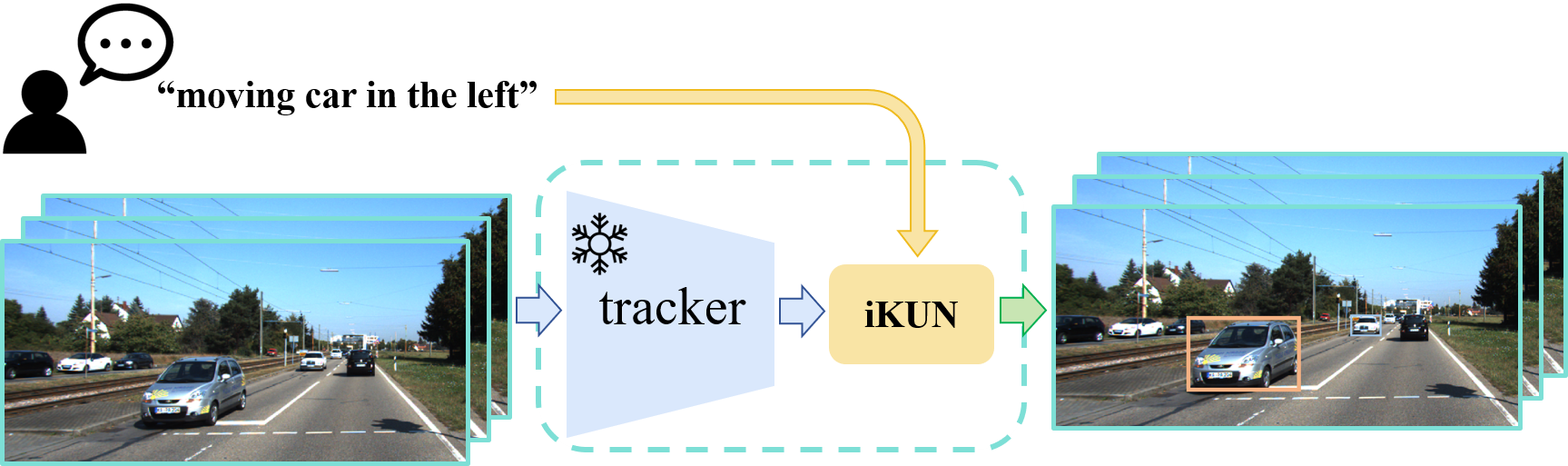}
    }
    \caption{
        \textbf{Comparison among previous RMOT frameworks and ours.}
        (a) Previous methods incorporate the referring module into the multi-object tracker,
        which need to retrain the overall framework.
        (b) Instead, our designed model iKUN can be directly plugged after an off-the-shelf tracker,
        in which the tracker is frozen while training.
    }
    \label{fig_1}
\end{figure}

\begin{figure*}
    \centering
    \subfloat[Two-stream framework without textual guidance.]{
        \includegraphics[width=.7\textwidth]{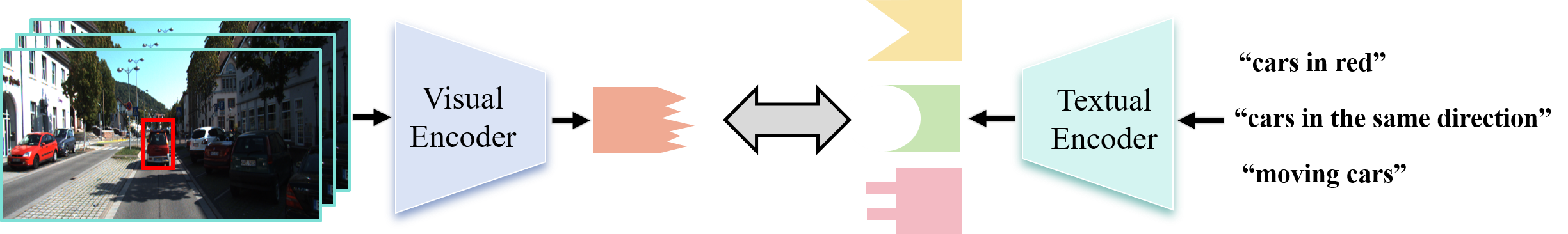}
    }\\
    \subfloat[Two-stream framework with textual guidance.]{
        \includegraphics[width=.7\textwidth]{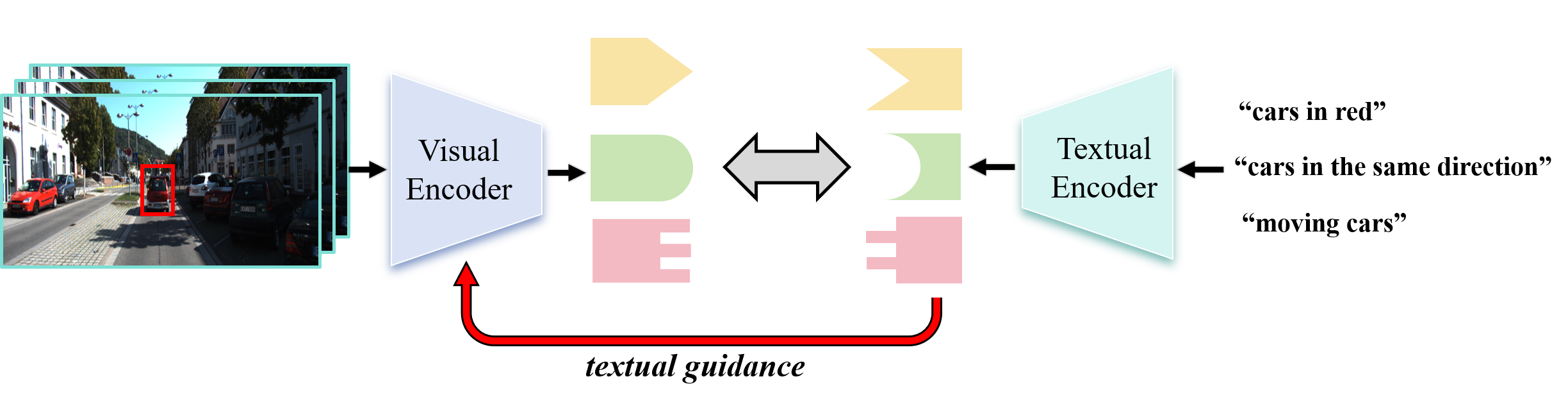}
    }
    \caption{
        \textbf{The motivation of KUM.}
        Given a tracklet and a set of descriptions,
        (a) without the guidance from textual stream, the visual encoder is asked to output a single feature to match multiple textual features;
        (b) with textual guidance, the visual encoder can predict adaptive features for each description.
    }
    \label{fig_2}
\end{figure*}

\section{Introduction}
    Traditional multi-object tracking (MOT) task aims to track all specific classes of objects frame-by-frame, which plays an essential role in video understanding.
    Although significant developments have been achieved, it suffers from poor flexibility and generalization.
    To address this problem, referring multi-object tracking (RMOT) task is recently proposed \cite{wu2023referring},
    whose core idea is to guide the multi-object tracking with a language description.
    For example, if we input ``moving car in the left'' as the query, the tracker will predict all trajectories corresponding to the description.
    However, as the cost of high flexibility, the model is required to perform detection, association and referring simultaneously.
    Therefore, balancing the optimization between subtasks becomes a critical issue.

    To accomplish this task, existing methods, e.g., TransRMOT \cite{wu2023referring}, 
    simply integrate a textual module into existing trackers, as shown in Fig.\ref{fig_1}(a).
    However, this framework has several intrinsic drawbacks:
    \textbf{i)} Task competition. 
    The optimization competition between detection and association has been revealed by some MOT methods \cite{liang2022rethinking, zhang2023motrv2}.
    In RMOT, the added referring subtask will further exacerbate this problem.
    \textbf{ii)} Engineering cost.
    Whenever we want to replace the baseline tracker, we need to rewrite the codes and retrain the entire framework.
    \textbf{iii)} Training cost. 
    Training all subtasks jointly results in high computational costs.

    Essentially, the tight bundling of tracking and referring subtasks is the main reason of these limitations.
    This raises a natural question: \textit{``Is it possible to decouple these two subtasks?''}
    In this work, we present a \textit{``tracking-to-referring''} framework with an insertable module \textbf{iKUN}, 
    which first tracks all candidates, and then recognizes the queried objects based on language descriptions.
    As shown in Fig.\ref{fig_1}(b), the tracker is frozen while training, and the optimization procedure can focus on the referring subtask.

    Therefore, the core problem lies in the design of an insertable referring module.
    An intuitive choice is the CLIP-style \cite{radford2021learning} module, which is pretrained on over 400 million image-text pairs for contrastive learning.
    Its main advantage is the excellent alignment of visual concepts and textual descriptions.
    For simplicity, the visual and textual streams of CLIP are independent.
    This means that for a given visual input, CLIP will extract a fixed visual feature regardless of the textual input.
    However, in the RMOT task, one trajectory often corresponds to multiple descriptions, including color, location, status, etc.
    It's hard to match a single feature with multiple various features.
    Motivated by this observation, we design a knowledge unification module (KUM) to adaptively extract visual features with the textual guidance.
    The illustration is shown in Fig.\ref{fig_2}.
    Moreover, to mitigate the effects of long-tail distribution of descriptions, we propose a test-time similarity calibration method to refine the referring results.
    The main idea is to estimate pseudo frequencies for descriptions in the open test set, and use them to revise the referring score.

    For the tracking subtask, Kalman filter \cite{kalman1960new} is widely used for motion modelling.
    The process noise and observation noise are two vital variables, which affect the accuracy of prediction and update steps.
    However, as a hand-crafted module, these two variables are determined by preset parameters, and are difficult to adapt to changes of motion status.
    We circumvent this problem by designing a neural version of Kalman filter, dubbed \textbf{NKF}, which dynamically estimates process and observation noises.

    We conduct extensive experiments on the recently released Refer-KITTI \cite{wu2023referring} dataset,
    and our methods show substantial superiority to existing solutions.
    Specifically, our solutions surpass previous SOTA method TransRMOT by \textbf{10.78\% HOTA}, \textbf{3.17\% MOTA} and \textbf{7.65\% IDF1}.
    Experiments for the traditional MOT task are also conducted on KITTI \cite{geiger2012we} and DanceTrack \cite{sun2022dancetrack},
    and the proposed NKF achieves noticeable improvement over baseline trackers \cite{wojke2017simple, zhang2022bytetrack}.
    To further verify the effectiveness of our methods, 
    we contribute a more challenging RMOT dataset, \textbf{Refer-Dance}, by extending DanceTrack with language descriptions.
    Our methods report a significant improvement than TransRMOT, i.e., 29.06\% vs. 9.58\% HOTA.

\begin{figure*}
    \centering
    \includegraphics[width=.75\textwidth]{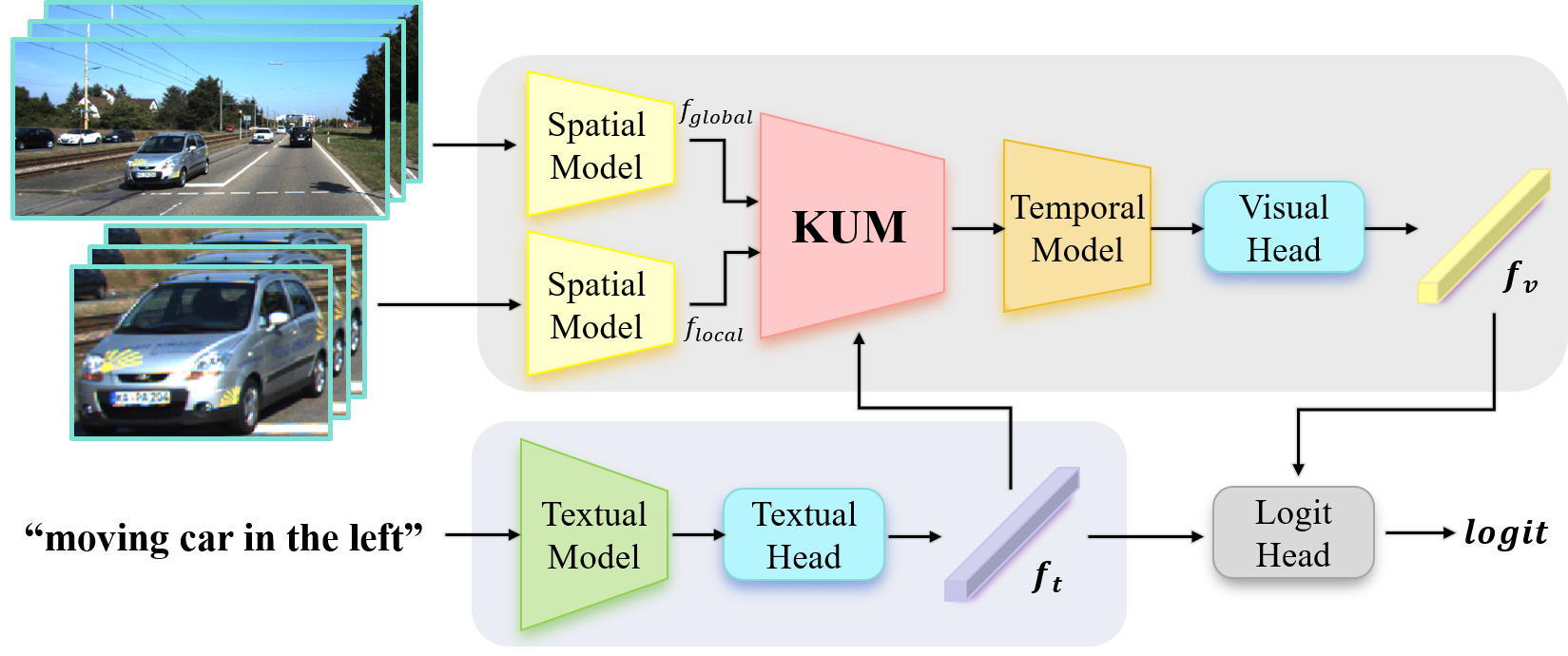}
    \caption{
        \textbf{The overall framework of iKUN.}
        The visual stream first embeds the local object feature $f_{local}$ and global scene feature $f_{global}$,
        and then aggregates them using the knowledge unification module (KUM).
        A temporal model and a visual head are followed to generate the final visual feature $f_v$.
        Meanwhile, the textual stream encodes the textual feature $f_t$.
        Finally, a logit head is utilized to predict the similarity score between $f_v$ and $f_t$
    }
    \label{fig_3}
\end{figure*}

\section{Related Work}
\subsection{Multi-object Tracking}
    Predominant approaches in MOT mainly follow the tracking-by-detection paradigm 
    \cite{yu2016poi, bochinski2017high, wang2020towards, zhang2021fairmot, yang2023hard}.
    They typically first predict the bounding boxes of objects and extract their features.
    Then an association step is used to match these instances across different frames.
    SORT \cite{bewley2016simple} applies Kalman filter for motion modelling and associates instances based on the intersection-over-union (IoU) of bounding boxes.
    DeepSORT \cite{wojke2017simple} extends it by adding an extra embedding network to extract appearance features of instances.
    OMC \cite{liang2022one} introduces an extra re-check network to restore missed targets.
    MAT \cite{han2022mat} focuses on high-performance motion-based prediction with a plug-and-play solution.
    ByteTrack \cite{zhang2022bytetrack} improves the association algorithm with the idea of ``associating every detection box''.
    Some recent SOTA methods, e.g., StrongSORT \cite{du2023strongsort}, OC-SORT \cite{cao2023observation} and BoT-SORT \cite{aharon2022bot},
    further design better association and post-processing solutions.
    
    Among these methods, the hand-crafted algorithm, Kalman filter, is widely used for motion modelling.
    However, it depends on elaborate parameter setting.
    To crack this nut, we introduce neural networks into Kalman filter, which can dynamically update the noise variables and adapt to more challenging scenarios.

\subsection{Referring Tracking}
    Referring single-object tracking (or segmentation) has been studied for several years.
    Benefiting from the flexibility of Transformer \cite{vaswani2017attention}, recent SOTA solutions mainly follow the joint tracking paradigm.
    TransVLT \cite{zhao2023transformer} designs proxy tokens to bridge between the visual and textual modalities, 
    which are followed by a Transformer-based cross-modal fusion module to estimate queried trajectories.
    JointNLT \cite{zhou2023joint} directly inputs language, template and text image embeddings into the Transformer encoder for relation modelling.
    MMTrack \cite{zheng2023towards} serializes language descriptions and bounding boxes into discrete tokens, and casts referring tracking as a token generation task.
    OVLM \cite{zhang2023one} proposes a one-stream network with memory token selection mechanism, which utilizes textual information to eliminate redundant tokens.
    MTTR \cite{botach2022end} applies a DETR-like \cite{carion2020end} multi-modal module to decode instance-level features into a set of multimodal sequences.
    ReferFormer \cite{wu2022language} inputs a set of object queries conditioned on language descriptions into Transformer to estimate the referred object.
    OnlineRefer \cite{wu2023onlinerefer} employs an elaborate query propagation mechanism to realize online referring tracking.

    Referring multi-object tracking is an emerging task, which can query an arbitrary number of instances \cite{wu2023referring}.
    Previous SOTA methods implement it by simply integrating the textual module to the tracking or detection model.
    Instead, we propose an insertable module to be plugged after any off-the-shelf trackers, which achieves much more flexibility and effectiveness.
    
\begin{figure*}
    \centering
    \subfloat[cascade attention]{
        \includegraphics[width=.3\textwidth]{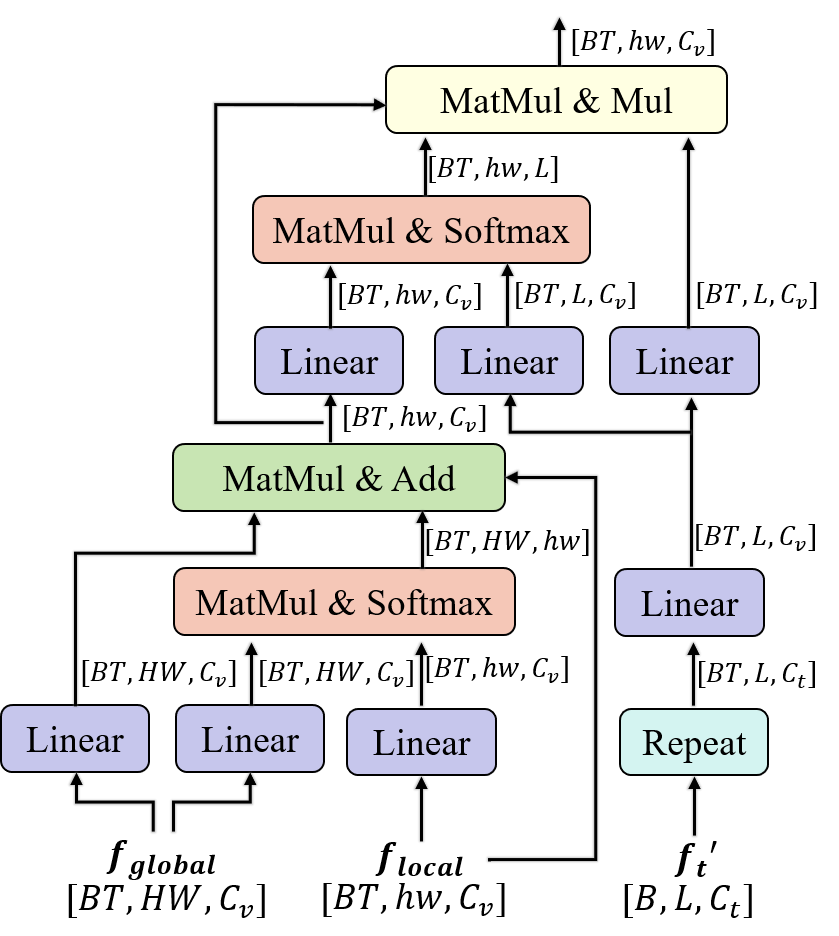}
    }
    \subfloat[cross correlation]{
        \includegraphics[width=.31\textwidth]{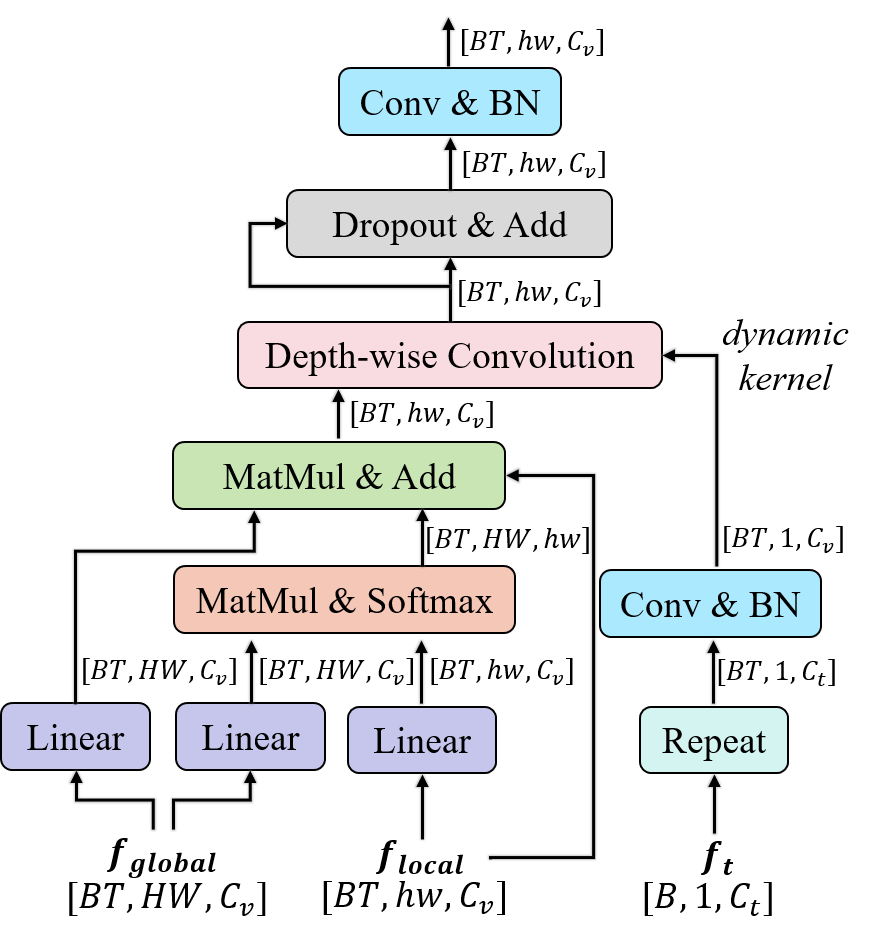}
    }
    \subfloat[text-first modulation]{
        \includegraphics[width=.22\textwidth]{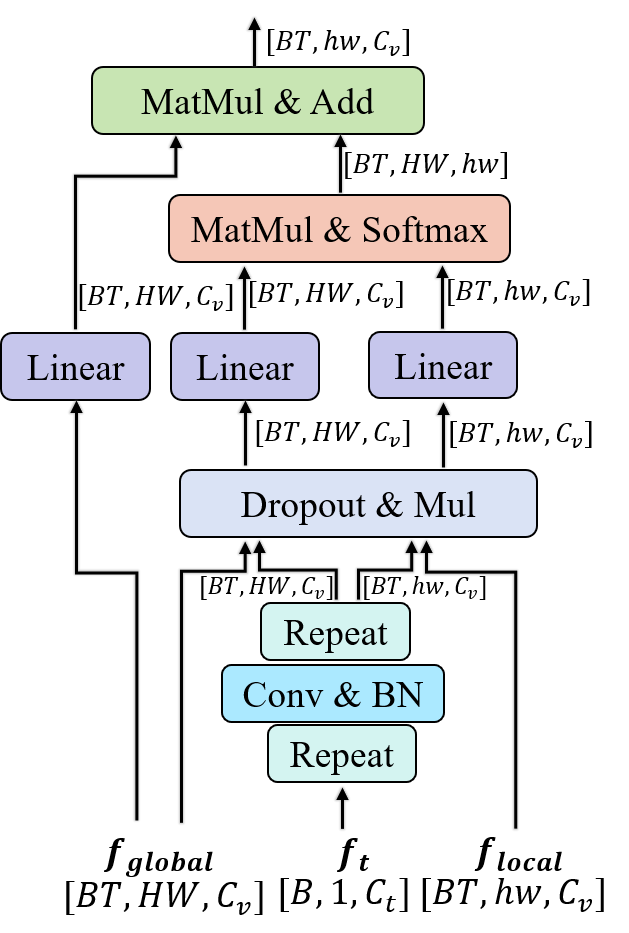}
    }
    \caption{
        \textbf{Three designs of knowledge unification module.}
        The feature maps are shown as the shape of their tensors with batch size $B$.
        For clarity, the final spatial global average pooling operation is omitted here.
    }
    \label{fig_4}
\end{figure*}

\section{Method}
\subsection{Method Overview}
    The input of RMOT consists of a sequence ${\cal I} = \{I_t\}_{k=1}^K$ with $K$ frames and a referring description ${\cal E} = \{e_l\}_{l=1}^L$ with $L$ words.
    Given an off-the-shelf tracker ${\cal F}_{trk}(\cdot)$, all candidate trajectories are predicted as ${\cal T} = {\cal F}_{trk}(\cal I)$,
    where ${\cal T} = \{ {\cal T}_1, {\cal T}_2, ..., {\cal T}_N\}$ includes $N$ instance trajectories. 
    Then, the referring module (i.e., iKUN) ${\cal F}_{ref}(\cdot)$ is applied to score candidates $\cal T$ by description as ${\cal S} = {\cal F}_{ref}(\cal I, T, E)$,
    which is further refined by the similarity calibration method.
    Finally, candidates ${\cal T}$ are filtered by refined scores $S'$ and the final outputs ${\cal T}'$ with $M \le N$ trajectories are generated.

    In the following sections, we first detail the design of ${\cal F}_{ref}$ in Sec.\ref{sec_ikun}. 
    Then the similarity calibration method is introduced in Sec.\ref{sec_sc}.
    Finally, we present the neural Kalman filter in Sec.\ref{sec_nkf} to improve the tracking performance of tracker ${\cal F}_{trk}$.

\subsection{Insertable Knowledge Unification Network}
\label{sec_ikun}
    iKUN is designed as a two-stream framework following CLIP \cite{radford2021learning} as shown in Fig.\ref{fig_3}.
    In the textual stream, description $\cal E$ with $L$ words is first encoded into $f_t' \in \mathbb R^{L \times C_t}$ 
    via a textual model $E_{txt}$, where $C_t$ is the dimension of textual feature.
    Then a textual head is utilized to squeeze $f_t'$ into $f_t \in \mathbb R^{C}$, where $C$ is the dimension of final feature. 

    For the visual stream, full images $\cal I$ and cropped images $\cal I'$ are encoded 
    into the global feature $f_{global} \in \mathbb R^{T \times HW \times C_v}$ 
    and local feature $f_{local} \in \mathbb R^{T \times hw \times C_v}$ with spatial model $E_{vis}$, respectively.
    Here, $T$ is temporal window size, $H, W, h, w$ represent spatial size of feature maps, and $C_v$ is the dimension of visual feature.
    Then the knowledge unification module (KUM) $E_{kum}$ is applied to aggregate them:
    \begin{equation}
        f_{uni} = E_{kum}(f_{global}, f_{local}) \in \mathbb R^{T \times C_v}. \label{eq1}
    \end{equation}    
    The temporal model and visual head are followed for temporal modelling and channel transformation, resulting in visual feature $f_v \in \mathbb R^{C}$.
    Finally, a logit head $H(\cdot)$ is utilized to compute the similarity score as $s = H(f_t, f_v)$.

    Though the two-stream framework is widely used in recent cross-modal retrieval methods 
    \cite{ma2022x, chun2021probabilistic, jiang2023cross, yan2023image, li2022learning, yan2022clip}, 
    we claim that it is not suitable for RMOT task.
    Specifically, given one trajectory, there may be multiple positive descriptions, e.g., ``cars in red'', ``cars in the same direction'' and ``moving cars''.
    However, the design of two independent streams is not sufficiently powerful for such ``one-to-many correspondence'' problem.

    To alleviate the above issues, we reformulate the knowledge unification module in Eq.\ref{eq1} as:
    \begin{equation}
        f_{uni} = E_{kum}(f_{global}, f_{local}, f_t) \in \mathbb R^{T \times C_v}. \label{eq2}
    \end{equation}
    That is, textual features are integrated into the unification process as guidance, which modulate visual feature extracting.
    In the following we will briefly introduce the three designs of KUM, as shown in Fig.\ref{fig_4}.
    The details are given in supplementary Sec.A.

    \noindent \textbf{Cascade attention.}
    The two visual features $f_{global}$ and $f_{local}$ are first aggregated via cross attention \cite{vaswani2017attention} with residual adding,
    where $f_{local}$ is query and $f_{global}$ is key / value.
    Then the resulting feature is fused with $f_t$ via another cross attention with residual multiplication.

    \noindent \textbf{Cross correlation.}
    Similarly, two visual features are first aggregated.
    Then a description conditioned dynamic convolutional operation is designed.
    Specifically, dynamic kernels are estimated based on textual feature $f_t$, and then are used to perform cross correlation with the aggregated visual feature.

    \noindent \textbf{Text-first Modulation.}
    The above two designs prioritize the visual aggregation.
    Instead, textual feature $f_t$ is introduced earlier here to modulate two visual features $f_{global}$ and $f_{local}$,
    which are later aggregated via cross attention and residual adding.

    In summary, ``cascade attention'' and ``cross correlation'' utilize different mechanisms to model the cross-modal relation,
    while in ``text-first modulation'', textual feature is used to guide visual aggregation.
    All these designs show substantial superiority to the baseline method (Eq.\ref{eq1}) in experiments, 
    which identifies the effectiveness of our solutions.


\begin{table*}[t]
    \renewcommand{\arraystretch}{1.2}
    \begin{center}
        \caption{
            \textbf{Comparison with state-of-the-art RMOT methods on Refer-KITTI.}
            $\dagger$: the results are from official code base \protect\footnotemark.
            $\star$: the similarity calibration method is applied.
            ``oracle'': the RMOT localization results are corrected based on GT.
            The results of first six rows are reported from TransRMOT \cite{wu2023referring}.
            Previous best results are bolded in \textcolor{blue}{blue}, and our best results are in \textcolor{red}{red}.
        }
        \label{table_rmot}
        \resizebox{.99\textwidth}{!}{
        \begin{tabular}{cl|c|c|c|c|c|c|c|c|c|c}
            \toprule[1pt]
            & \textbf{Method} & \textbf{Detector} & \textbf{HOTA} & \textbf{DetA} & \textbf{AssA} & \textbf{DetRe} & \textbf{DetPr} & \textbf{AssRe} & \textbf{AssPr} & \textbf{MOTA} & \textbf{IDF1}\\
            \hline
            & FairMOT$^\dagger$   \cite{zhang2021fairmot}         & DLA-34 \cite{yu2018deep}                      & 23.46 & 14.84 & 40.15 & 17.40 & 43.58 & 53.35 & 73.15 & 0.80  & 26.18 \\
            & DeepSORT            \cite{wojke2017simple}          & DLA-34 \cite{yu2018deep}                      & 25.59 & 19.76 & 34.31 & 26.38 & 36.93 & 39.55 & 61.05 &   -   &   -   \\
            & ByteTrack$^\dagger$ \cite{zhang2022bytetrack}       & DLA-34 \cite{yu2018deep}                      & 22.49 & 13.17 & 40.62 & 16.13 & 36.61 & 46.09 & 73.39 & -7.52 & 23.72 \\
            & CSTrack             \cite{liang2022rethinking}      & YOLOv5 \cite{jocher2022ultralytics}           & 27.91 & 20.65 & 39.10 & 33.76 & 32.61 & 43.12 & 71.82 &   -   &   -   \\
            & TransTrack          \cite{sun2020transtrack}        & DeformableDETR \cite{zhu2020deformable}       & 32.77 & 23.31 & 45.71 & 32.33 & 42.23 & 49.99 & 78.74 &   -   &   -   \\
            & TrackFormer         \cite{meinhardt2022trackformer} & DeformableDETR \cite{zhu2020deformable}       & 33.26 & 25.44 & 45.87 & 35.21 & 42.19 & 50.26 & 78.92 &   -   &   -   \\
            & TransRMOT$^\dagger$ \cite{wu2023referring}          & DeformableDETR \cite{zhu2020deformable}       & \textcolor{blue}{\textbf{38.06}} & \textcolor{blue}{\textbf{29.28}}
            & \textcolor{blue}{\textbf{50.83}} & \textcolor{blue}{\textbf{40.20}} & \textcolor{blue}{\textbf{47.36}} & \textcolor{blue}{\textbf{55.43}} 
            & \textcolor{blue}{\textbf{81.36}} & \textcolor{blue}{\textbf{9.03}}  & \textcolor{blue}{\textbf{46.40}} \\
            \hline
            & ByteTrack\cite{zhang2022bytetrack}+iKUN$^\star$     & YOLOv8 \cite{Jocher_YOLO_by_Ultralytics_2023} & 41.25 & 29.59 & 57.83 & 44.23 & 43.39 & 63.77 & 74.96 & 5.27  & 51.82 \\ 
            & OC-SORT\cite{cao2023observation}+iKUN$^\star$       & YOLOv8 \cite{Jocher_YOLO_by_Ultralytics_2023} & 42.08 & 29.76 & 60.01 & 42.39 & 46.30 & 64.76 & 81.26 & 11.02 & 53.16 \\
            & DeepSORT\cite{wojke2017simple}+iKUN$^\star$         & YOLOv8 \cite{Jocher_YOLO_by_Ultralytics_2023} & 42.46 & 31.64 & 57.56 & 46.03 & 46.32 & 63.48 & 77.66 & \textcolor{red}{\textbf{12.50}} & 52.57 \\
            & Deep OC-SORT\cite{maggiolino2023deep}+iKUN$^\star$  & YOLOv8 \cite{Jocher_YOLO_by_Ultralytics_2023} & 42.94 & 29.90 & 62.15 & 46.35 & 42.45 & 69.42 & 77.99 & 3.63  & 53.71 \\ 
            & StrongSORT\cite{du2023strongsort}+iKUN$^\star$      & YOLOv8 \cite{Jocher_YOLO_by_Ultralytics_2023} & 43.30 & 31.44 & 60.09 & 47.14 & 44.33 & 66.60 & 76.25 & 10.10 & 54.05 \\ 
            \hline
            & \textbf{NeuralSORT+iKUN$^\star$ (ours)}             & \textbf{YOLOv8 \cite{Jocher_YOLO_by_Ultralytics_2023}} 
            & \textbf{44.56} & \textbf{32.05} & \textbf{62.48} & \textbf{48.53} & \textbf{44.76} & \textbf{70.52} & \textbf{76.66} & \textbf{9.69}  & \textcolor{red}{\textbf{55.40}} \\
            &                                                     & \textbf{DeformableDETR \cite{zhu2020deformable}}       
            & \textcolor{red}{\textbf{48.84}} & \textcolor{red}{\textbf{35.74}} & \textcolor{red}{\textbf{66.80}} & \textcolor{red}{\textbf{51.97}} 
            & \textcolor{red}{\textbf{52.25}} & \textcolor{red}{\textbf{72.95}} & \textcolor{red}{\textbf{87.09}} & \textbf{12.26} & \textbf{54.05} \\
            \hline
            & FairMOT$^\dagger$   \cite{zhang2021fairmot}         & oracle                                        & 33.02 & 26.39 & 41.33 & 28.13 & 81.03 & 42.54 & 88.58 & 18.72 & 33.46 \\
            & ByteTrack$^\dagger$ \cite{zhang2022bytetrack}       & oracle                                        & 35.23 & 25.69 & 48.30 & 27.82 & 77.03 & 51.12 & 86.45 & 18.26 & 35.29 \\
            & TransRMOT$^\dagger$ \cite{wu2023referring}          & oracle                                        & 54.50 & 48.74 & 60.95 & 56.67 & 77.70 & 63.23 & 93.82 & 39.44 & 58.57 \\
            & \textbf{NeuralSORT+iKUN$^\star$ (ours)}             & \textbf{oracle}                               & \textbf{61.54} & \textbf{48.59} & \textbf{77.94} 
                                                                  & \textbf{64.06} & \textbf{66.79} & \textbf{82.92} & \textbf{91.23} & \textbf{31.84} & \textbf{62.05} \\
            \bottomrule[1pt]
        \end{tabular}
        }
    \end{center}
  \end{table*}

\subsection{Similarity Calibration}
\label{sec_sc}
    Learning under long-tail distribution has been widely studied in recent years \cite{cui2019class, hyun2022long, zhao2023mdcs, du2023superdisco, du2023no}.
    The similar problem is also observed in RMOT task, that is, 
    there is a huge difference in the number of positive instances of descriptions.  
    However, it has the following two significant differences from the traditional long-tail distribution problem:
    \textbf{i)} \textbf{Non-uniform test set.}
    Most previous works assume the long-tail distribution on training set and uniform distribution on test set.
    However, the test set of RMOT datasets follows non-uniform distribution, which makes the existing solutions ineffective.
    \textbf{ii)} \textbf{Open test set.} 
    RMOT is an open-set task, i.e., there are unseen descriptions during the test time.
    That means previous training-time solutions may not work on the test set.

    Inspired by above observations, we propose a test-time similarity calibration method to refine the predicted similarity score by iKUN.
    Specifically, the frequencies of all $N_{tr}$ training descriptions $\{ {\cal E}_i^{tr} \}_{i=1}^{N_{tr}}$ are calculated, denoted as $\{ p_i^{tr} \}_{i=1}^{N_{tr}}$.
    Given a test description ${\cal E}_j^{ts}$, the normalized similarity between it and ${\cal E}_i^{tr}$ is formulated as:
    \begin{equation}
        w_{ij} = {exp(\tau \cdot x_{ij}) \over {\sum_k exp(\tau \cdot x_{ik})}}, \label{eq3}
    \end{equation}
    where $\tau$ (set to 100) is the temperature parameter, and $x_{ij}$ is the similarity score estimated by any language model.
    Then, the \textit{pseudo frequency} of ${\cal E}_j^{ts}$ can be estimated by $p_j^{ts} = \sum_i w_{ij} \cdot p_i^{tr}$,
    which is further utilized to refine the similarity score $s_j$ from iKUN as $s_j' = s_j + f(p_j^{ts})$.
    Here, $f(\cdot)$ is designed as a linear function $f(x) = a \cdot x + b$ for simplicity.

\footnotetext{https://github.com/wudongming97/RMOT}

\subsection{Neural Kalman Filter}
\label{sec_nkf}
    Kalman filter \cite{kalman1960new} is widely used to estimate motion status of tracked objects in MOT.
    It operates in two distinct phases, i.e., state prediction and state update,
    in which Kalman gain is designed to balance the weights of estimates and observations.
    Concretely, Kalman gain $K_k$ at time step $k$ is calculated as:
    \begin{equation}
        K_k = P_k' H_k^T (H_k P_k' H_k^T + R_k)^{-1}, \label{eq4}
    \end{equation}
    where $H_k$ is observation model, $R_k$ is observation noise, and $P_k'$ is the predicted covariance by:
    \begin{equation}
        P_k' = F_k P_{k-1} F_k^T + Q_k, \label{eq5}
    \end{equation}
    where $F_k$ is state transition model, $Q_k$ is process noise, and $P_{k-1}$ is the state covariance at time step $k-1$.

    It can be observed that $K_k$ is greatly influenced by $R_k$ and $Q_k$.
    However, in previous multi-object trackers, they are typically determined by preset parameters, which are hard to adapt to various scenarios.
    Inspired by it, we construct two neural networks, i.e., R-Net $F_R$ and Q-Net $F_Q$, to dynamically update $R_k$ and $Q_k$ based on the current motion status:
    \begin{equation}
        R_k = F_R (z_k),\ Q_k = F_Q (x_{k-1}),
    \end{equation}
    where $z_k$ is the current observations, and $x_{k-1}$ is the state mean at time step $k-1$.

    In implementation, $F_R$ and $F_Q$ are designed as fully connected layers.
    More complex structures (e.g., LSTM \cite{hochreiter1997long} and GRU \cite{cho2014learning}) are also tried, but don't show obvious improvements.

\begin{table*}
    \renewcommand{\arraystretch}{1.2}
    \begin{center}
        \caption{
            \textbf{Comparison with state-of-the-art MOT methods on KITTI.}
            All trackers use the same detection results from YOLOv8.
        }
        \label{table_mot}
        \resizebox{.9\textwidth}{!}{
        \begin{tabular}{cl|c|c|c|c|c|c|c|c|c|c}
            \toprule[1pt]
            & \multirow{2}*{\textbf{Method}} & \multicolumn{5}{c|}{\textbf{Car}} & \multicolumn{5}{c}{\textbf{Pedestrian}} \\
            \cline{3-12}
            & ~ & \textbf{HOTA} & \textbf{DetA} & \textbf{AssA} & \textbf{MOTA} & \textbf{IDF1} & \textbf{HOTA} & \textbf{DetA} & \textbf{AssA} & \textbf{MOTA} & \textbf{IDF1} \\
            \hline
            & ByteTrack           \cite{zhang2022bytetrack} & 56.34 & 51.15 & 62.58 & 57.29 & 75.24 & 38.87 & 33.84 & 45.35 & 23.16 & 57.85 \\
            & OC-SORT             \cite{cao2023observation} & 57.24 & 51.24 & 64.60 & 60.23 & 76.54 & 40.35 & 35.24 & 46.54 & 30.41 & 60.46 \\
            & Deep OC-SORT        \cite{maggiolino2023deep} & 58.78 & 52.85 & 66.08 & 60.01 & 78.12 & 45.65 & 38.22 & 54.77 & 30.60 & 66.61 \\
            & StrongSORT          \cite{du2023strongsort}   & 59.78 & 55.69 & 64.96 & 65.36 & 77.78 & 48.89 & 43.22 & 55.61 & 51.60 & 73.58 \\
            \hline
            & DeepSORT (baseline) \cite{wojke2017simple}    & 58.58 & 56.55 & 61.51 & 66.99 & 76.00 & 45.03 & 39.37 & 51.93 & 46.33 & 68.97 \\
            & \textbf{NeuralSORT (ours)}                    & \textbf{61.48} & \textbf{57.18} & \textbf{66.91} & \textbf{66.60} & \textbf{80.33} 
                                                            & \textbf{50.47} & \textbf{44.05} & \textbf{58.19} & \textbf{52.35} & \textbf{74.47} \\
            \bottomrule[1pt]
        \end{tabular}
        }
    \end{center}
\end{table*}

\begin{figure}
    \centering
    \includegraphics[width=.45\textwidth]{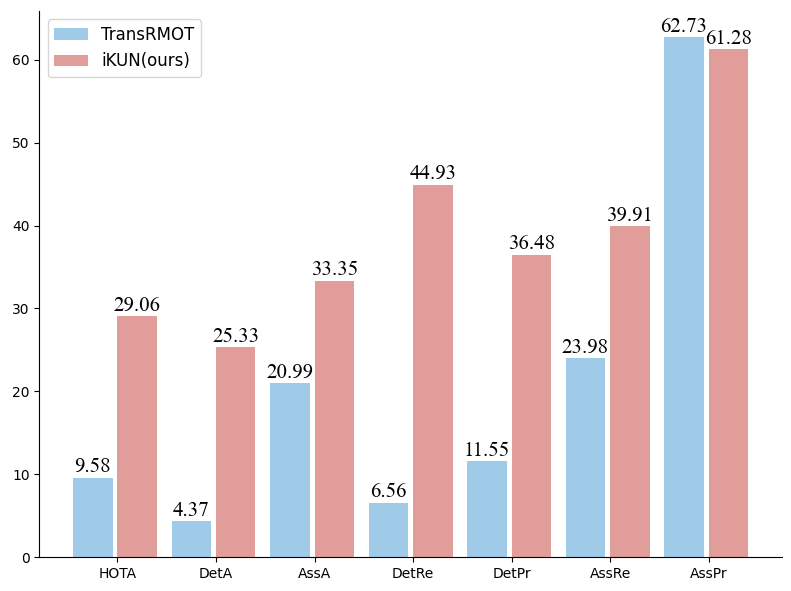}
    \caption{
        \textbf{The performance of TransRMOT\cite{wu2023referring} and iKUN on Refer-Dance.}
    }
    \label{fig_6}
\end{figure}

\section{Experiment}

\subsection{Experimental setup}
    \noindent\textbf{Dataset and Metrics.}
    Refer-KITTI \cite{wu2023referring} is currently the only public dataset for referring multi-object tracking,
    which is extended from KITTI \cite{geiger2012we}.
    We follow the official data split protocols, in which 15 videos with 80 distinct descriptions are used for training,
    and 3 videos with 63 distinct descriptions are used for testing.
    HOTA series \cite{luiten2021hota}, MOTA \cite{bernardin2008evaluating}, IDF1 \cite{ristani2016performance} are used for evaluation.

    We also construct a more challenging RMOT dataset Refer-Dance by extending DanceTrack \cite{sun2022dancetrack}.
    It contains 40 videos with 39 distinct descriptions for training, and 25 videos with 17 distinct descriptions for testing.
    The description annotations focus on motion and dressing status, 
    e.g., ``dancing person with black T-shirt and green pants'' and ``standing person dressed all in black''.
    Please refer to supplementary Sec.B for more examples.

    To validate the effectiveness of proposed neural Klaman filter, we further conduct experiments on KITTI and DanceTrack.
    The same evaluation metrics as above are utilized.

    \noindent\textbf{Implementation Details.}
    For iKUN, we borrow the visual and textual encoders from \texttt{CLIP-RN50} \cite{radford2021learning}.
    The feature dimensions are set as $C_v = 2048$, $C_t = C = 1024$.
    The window size $T$ is set to 8 with a stride of 4.
    The visual and textual heads are multi-layer perceptrons, temporal model is realized by temporal average pooling, and logit head is realized by cosine similarity.
    The model is trained on ground truth tracklets with focal loss \cite{lin2017focal} for 100 epochs.
    The initial learning rate is set to 1e-5 and decays according to the cosine annealing strategy.
    The textual model is frozen for training stability.
    For similarity calibration, pretrained CLIP is applied for textual encoding.
    The parameters in mapping function $f(\cdot)$ are set as $a=8, b=-0.1$.
    
    For neural Kalman filter (NKF), R-Net and Q-Net are jointly trained for 10 epochs with mean square error loss.
    The learning rate is set as the same as iKUN.
    For inference, we integrate NKF into DeepSORT \cite{wojke2017simple}, along with the following tricks:
    i) Extra exiting decision. Trajectories whose estimated positions exceed the image range are deleted.
    ii) Extra velocity cost. Extra association cost term based on velocity is introduced inspired by OC-SORT \cite{cao2023observation}.
    iii) Linear interpolation. Missing detections are restored by linear interpolation.
    The final tracker is termed as \textbf{NeuralSORT}.

\begin{table*}
    \begin{center}
        \caption{
            \textbf{Ablation study on different designs of knowledge unification module.}
            ``YOLOv8+NeuralSORT'' are used as multi-object tracker.
            The default setting is marked in \colorbox{lightgray}{gray}.
        }
        \label{table_kun}
        \resizebox{.75\textwidth}{!}{
        \begin{tabular}{cl|c|c|c|c|c|c|c|c}
            \toprule[1pt]
            & \textbf{Method} & \textbf{KUM} & \textbf{HOTA} & \textbf{DetA} & \textbf{AssA} & \textbf{DetRe} & \textbf{DetPr} & \textbf{AssRe} & \textbf{AssPr} \\
            \hline
            & baseline & -                          & 39.09 & 26.13 & 58.90 & 33.12 & 50.33 & 66.93 & 75.51 \\
            &          & \cellcolor{lightgray}cascade attention & \cellcolor{lightgray}43.75 & \cellcolor{lightgray}31.96 & \cellcolor{lightgray}60.39 
                                   & \cellcolor{lightgray}44.70 & \cellcolor{lightgray}48.39 & \cellcolor{lightgray}68.34 & \cellcolor{lightgray}76.64 \\
            &          & cross correlation          & 40.14 & 28.30 & 57.65 & 40.18 & 45.10 & 65.32 & 76.92 \\
            &          & text-first modulation      & 40.90 & 26.58 & 63.62 & 53.93 & 32.50 & 72.66 & 76.73 \\
            \bottomrule[1pt]
        \end{tabular}
        }
    \end{center}
\end{table*}

\begin{table*}
    \begin{center}
        \caption{
            \textbf{Ablation study on different components of NeuralSORT on KITTI}.
            NKF: neural kalman filter;
            DEL: extra exiting decision;
            VEL: extra velocity cost;
            INT: linear interpolation.
            The default setting is marked in \colorbox{lightgray}{gray}.
        }
        \label{table_neuralsort}
        \resizebox{.75\textwidth}{!}{
        \begin{tabular}{cl|c|c|c|c|c|c|c|c|c|c}
            \toprule[1pt]
            & \multirow{2}*{\textbf{Method}} & \multirow{2}*{\textbf{NKF}} & \multirow{2}*{\textbf{DEL}} & \multirow{2}*{\textbf{VEL}} & \multirow{2}*{\textbf{INT}} 
                                                & \multicolumn{3}{c|}{\textbf{Car}} & \multicolumn{3}{c}{\textbf{Pedestrian}} \\
            \cline{7-12}
            & ~ & ~ & ~ & ~ & ~ & \textbf{HOTA} & \textbf{DetA} & \textbf{AssA} & \textbf{HOTA} & \textbf{DetA} & \textbf{AssA} \\
            \hline
            & DeepSORT  \cite{maggiolino2023deep} &      -     &      -     &      -     &      -     & 58.58 & 56.55 & 61.51 & 45.03 & 39.37 & 51.93 \\
            &                                     & \checkmark &            &            &            & 59.90 & 56.85 & 63.91 & 48.53 & 42.40 & 55.88 \\
            &                                     &            & \checkmark &            &            & 58.69 & 56.55 & 61.74 & 45.03 & 39.37 & 51.93 \\
            &                                     &            &            & \checkmark &            & 58.81 & 56.65 & 61.90 & 47.25 & 42.48 & 52.94 \\
            &                                     &            &            &            & \checkmark & 57.64 & 54.63 & 61.68 & 45.86 & 39.89 & 53.15 \\
            &                                     & \checkmark & \checkmark &            &            & 60.61 & 56.87 & 65.40 & 48.53 & 42.40 & 55.88 \\
            &                                     & \checkmark & \checkmark & \checkmark &            & 61.19 & 57.00 & 66.49 & 49.15 & 43.79 & 55.51 \\
            & \cellcolor{lightgray}NeuralSORT     & \cellcolor{lightgray}\checkmark & \cellcolor{lightgray}\checkmark & \cellcolor{lightgray}\checkmark 
                                                  & \cellcolor{lightgray}\checkmark & \cellcolor{lightgray}61.48 & \cellcolor{lightgray}57.18 
                                                  & \cellcolor{lightgray}66.91 & \cellcolor{lightgray}50.47 & \cellcolor{lightgray}44.05 & \cellcolor{lightgray}58.19 \\
            \bottomrule[1pt]
        \end{tabular}
        }
    \end{center}
\end{table*}

\begin{table}
    \begin{center}
        \caption{
            \textbf{Effect of hyper-parameters of similarity calibration.}
            The second line ``cascade attention'' in Tab.\ref{table_kun} are taken as baseline.
            The selected parameters are marked in \colorbox{lightgray}{gray}.
        }
        \label{table_sc}
        \resizebox{.45\textwidth}{!}{
        \begin{tabular}{cl|c|c|c|c|c|c}
            \toprule[1pt]
            & \textbf{a} & \textbf{b} & \textbf{HOTA} & \textbf{DetA} & \textbf{AssA} & \textbf{DetRe} & \textbf{AssRe} \\
            \hline
            & \underline{0} & \underline{0} & \underline{43.75} & \underline{31.96} & \underline{60.39} & \underline{44.70} & \underline{68.34} \\
            & 0 & -0.1 & 43.06 & 31.21 & 59.82 & 41.43 & 67.67 \\
            & 0 & -0.2 & 41.80 & 29.82 & 59.00 & 37.88 & 66.87 \\
            & 2 &   0  & 44.20 & 32.18 & 61.26 & 46.49 & 69.25 \\
            & 2 & -0.1 & 43.66 & 31.92 & 60.19 & 43.63 & 68.09 \\
            & 2 & -0.2 & 42.80 & 30.92 & 59.65 & 40.42 & 67.45 \\
            & 4 &   0  & 44.36 & 32.08 & 61.89 & 48.06 & 69.95 \\
            & 4 & -0.1 & 44.07 & 32.14 & 60.99 & 45.39 & 68.94 \\
            & 4 & -0.2 & 43.55 & 31.72 & 60.24 & 42.58 & 68.08 \\
            & 8 &   0  & 44.10 & 31.10 & 63.11 & 50.41 & 71.35 \\
            & \cellcolor{lightgray}8 & \cellcolor{lightgray}-0.1 & \cellcolor{lightgray}44.56 & \cellcolor{lightgray}32.05 & \cellcolor{lightgray}62.48 
                                                                    & \cellcolor{lightgray}48.53 & \cellcolor{lightgray}70.52 \\
            & 8 & -0.2 & 44.22 & 32.09 & 61.47 & 45.91 & 69.42 \\
            \bottomrule[1pt]
        \end{tabular}
        }
    \end{center}
\end{table}

\subsection{Benchmark Experiments}
    \noindent\textbf{Refer-KITTI.}
    We compare our methods with previous solutions on Refer-KITTI in Tab.\ref{table_rmot}.
    Current SOTA method, TransRMOT \cite{wu2023referring}, obtains 38.06\%, 29.28\%, 50.83\%, corresponding to HOTA, DetA, AssA, respectively.
    In comparison, we integrate our iKUN into various off-the-shelf trackers based on YOLOv8 \cite{Jocher_YOLO_by_Ultralytics_2023},
    and achieve consistent improvements, i.e., 41.25\%-44.56\% HOTA.
    By switching to the same detector as TransRMOT, i.e., DeformableDETR \cite{zhu2020deformable}, 
    we achieve 48.84\%, 35.74\%, 66.80\%, corresponding to HOTA, DetA, AssA, respectively.
    Importantly, benefiting from the flexibility of our framework, iKUN is only trained once for multiple trackers.

    Moreover, to focus on the comparison of the association and referring ability, 
    we conduct oracle experiments to eliminate the interference of localization accuracy.
    That is, the coordinates $(x,y,w,h)$ of final estimated trajectories are revised based on ground truth.
    Please note that no bounding boxes are added or deleted, and no IDs are modified.
    In this setting, our methods are also performant compared with TransRMOT, i.e., 61.54\% vs. 54.50\% HOTA.

    \noindent\textbf{Refer-Dance.}
    We further compare our methods with TransRMOT on our constructed Refer-Dance dataset.
    ByteTrack \cite{zhang2022bytetrack} with NKF is taken as our baseline tracker.
    The comparison results are shown in Fig.\ref{fig_6}, 
    It can be observed that our methods achieve much better results among main metrics, 
    e.g., 29.06\% vs. 9.58\% HOTA, 25.33\% vs. 4.37\% DetA and 33.35\% vs. 20.99\% AssA.

    \noindent\textbf{KITTI.}
    We compare the designed NeuralSORT with current SOTA trackers on KITTI in Tab.\ref{table_mot}.
    All trackers utilize the same detections from YOLOv8.
    For simplicity, we use the same data split protocol as in Refer-KITTI.
    It is shown that our NeuralSORT achieves the best results for both car and pedestrian classes.

\subsection{Ablation Experiments}
    \noindent\textbf{Knowledge unification module.}
    The three designs of KUM are compared in Tab.\ref{table_kun}.
    Unification without textual guidance in Eq.\ref{eq1} is taken as baseline.
    It is shown that all these strategies can bring remarkable improvements over baseline method,
    which demonstrates the effectiveness of the textual guidance mechanism.
    In detail, ``text-first modulation'' achieves best association performance (AssA), but is poor in detection (DetA).
    ``Cross correlation'' obtains higher DetA but lower AssA.
    ``Cascade attention'' achieves the best results for HOTA and DetA metrics, and is comparable for AssA metrics.
    Finally, we choose ``cascade attention'' as the default design of KUM.

\begin{table}
    \renewcommand{\arraystretch}{1.2}
    \begin{center}
        \caption{
            \textbf{The comparison between vanilla Kalman filter and neural Kalman filter for multiple pedestrian tracking.}
            We use ByteTrack\cite{zhang2022bytetrack} as the baseline tracker and experiment on the KITTI and DanceTrack dataset.
        }
        \label{table_nkf}
        \resizebox{.47\textwidth}{!}{
        \begin{tabular}{cl|c|c|c|c|c|c}
            \toprule[1pt]
            & \textbf{Dataset} & \textbf{Kalman} & \textbf{HOTA} & \textbf{DetA} & \textbf{AssA} & \textbf{MOTA} & \textbf{IDF1} \\
            \hline
            &   KITTI    & vanilla & 38.87 & 33.84 & 45.35 & 23.16 & 57.85 \\
            &            & \textbf{neural}  & \textbf{44.89} & \textbf{37.54} & \textbf{54.08} & \textbf{32.02} & \textbf{64.46} \\
            \hline
            & DanceTrack & vanilla & 46.88 & 70.18 & 31.44 & 87.56 & 52.32 \\
            &            & \textbf{neural}  & \textbf{54.52} & \textbf{78.14} & \textbf{38.19} & \textbf{89.37} & \textbf{54.67} \\
            \bottomrule[1pt]
        \end{tabular}
        }
    \end{center}
\end{table}

\begin{table}
    \begin{center}
        \caption{
            \textbf{Comparison of training and inference time between TransRMOT and iKUN.}
            Both models are trained for 100 epochs.
            All experiments are conducted on the same machine with multiple Tesla T4 GPUs on dataset Refer-KITTI.
        }
        \label{table_time}
        \resizebox{.45\textwidth}{!}{
        \begin{tabular}{cl|c|c|c}
            \toprule[1pt]
            & \textbf{Method} & \textbf{\ \ \ \ \ mode\ \ \ \ \ } & \textbf{GPU number} & \textbf{time cost} \\
            \hline
            &   TransRMOT & training  & 4 & 44 hours 53 minutes \\ 
            &             & inference & 1 & 2 hours 34 minutes \\
            \hline
            &   iKUN      & training  & 2 & 2 hours 25 minutes \\
            &             & inference & 1 & 1 hour 38 minutes \\
            \bottomrule[1pt]
        \end{tabular}
        }
    \end{center}
\end{table}  

    \noindent\textbf{Similarity calibration.}
    We investigate the effect of hyper-parameters $a, b$ in the mapping function $f(\cdot)$ in Tab.\ref{table_sc}.
    As reported, the performance is robust to the varying values.
    In this work, we choose $a = 8$ and $b = -0.1$ as default,
    which achieves performance gains of 0.81\% HOTA and 2.09\% AssA.
 
    \noindent\textbf{Neural Kalman Filter.}
    We first take DeepSORT as baseline and study different components of NeuralSORT on KITTI in Tab.\ref{table_neuralsort}.
    Most importantly, NKF improves HOTA by 1.32\% for car and 3.50\% for pedestrian.
    Other tricks further bring gains of 1.58\% and 1.94\% for car and pedestrian respectively.
    Then, we take ByteTrack as baseline and further investigate the effect of NKF on KITTI and DanceTrack in Tab.\ref{table_nkf}.
    Significant improvements can be observed on both two datasets for all evaluation metrics,
    which shows the superiority of our method.

    \noindent\textbf{Training and inference time.}
    One concern is the computational cost of our multi-stage framework.
    We conduct experiments on Refer-KITTI with multiple Tesla T4 GPUs and compare the training and inference time between TransRMOT and iKUN in Tab.\ref{table_time}.
    It can be observed that our method achieves much lower time cost.
    Note that, for fair comparison, the tracking process is also included for the inference time.

\begin{figure}[h]
    \centering
    \includegraphics[width=.47\textwidth]{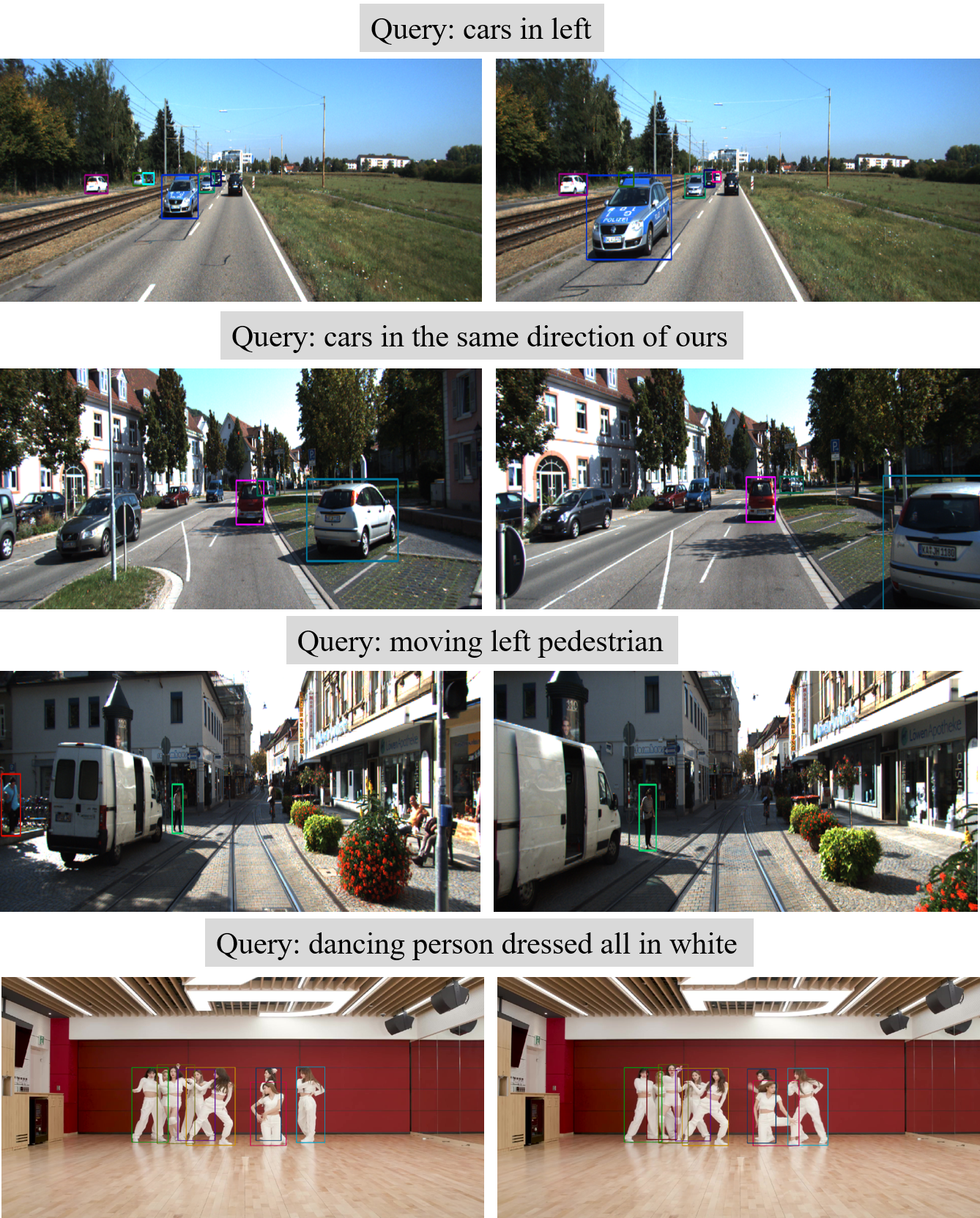}
    \caption{
        \textbf{Qualitative results of our method on Refer-KITTI (row 1 to 3) and Refer-Dance (row 4).}
    }
    \label{fig_7}
\end{figure}

\subsection{Qualitative Results}
    We visualize several typical referring results in Fig.\ref{fig_7}.
    The first query focuses on the location of cars, the second one describes the orientation of cars, the third query includes the motion and location of persons,
    and the fourth one attends to the motion and dressing status of persons.
    Our methods can successfully track targets based on various queries.

\subsection{Limitations}
    The main limitation of our multi-stage framework is the miscellaneous engineering details.
    For example, there exist several nontrivial hyperparameters that need to be tuned in the tracking, referring and post-processing components.
    Moreover, the temporal modelling capability of our model is limited, which constrains the performance for motion-related queries.

\section{Conclusion}
    In this work, we present a novel module, iKUN, which can be plugged after any multi-object trackers to realize referring tracking.
    To address the one-to-many correspondence problem, knowledge unification module is designed to modulate visual embeddings based on textual descriptions.
    The similarity calibration method is further proposed to refine predicted scores with estimated pseudo frequencies in the open test set.
    Moreover, two light-weight neural networks are introduced into Kalman filter to dynamically update the process and observation noise variables.
    The effectiveness of our methods is demonstrated by experiments on the public dataset Refer-KITTI and our newly constructed dataset Refer-Dance.

\section*{Acknowledgments}
    This work is supported by Chinese National Natural Science Foundation under Grants (62076033)
    and BUPT Excellent Ph.D. Students Foundation (CX2022145).

{
    \small
    \bibliographystyle{ieeenat_fullname}
    \bibliography{main}
}
\end{document}

%% file: preamble.tex
\usepackage[dvipsnames]{xcolor}


%% file: main.bbl
\begin{thebibliography}{53}
\providecommand{\natexlab}[1]{#1}
\providecommand{\url}[1]{\texttt{#1}}
\expandafter\ifx\csname urlstyle\endcsname\relax
  \providecommand{\doi}[1]{doi: #1}\else
  \providecommand{\doi}{doi: \begingroup \urlstyle{rm}\Url}\fi

\bibitem[Aharon et~al.(2022)Aharon, Orfaig, and Bobrovsky]{aharon2022bot}
Nir Aharon, Roy Orfaig, and Ben-Zion Bobrovsky.
\newblock Bot-sort: Robust associations multi-pedestrian tracking.
\newblock \emph{arXiv preprint arXiv:2206.14651}, 2022.

\bibitem[Bernardin and Stiefelhagen(2008)]{bernardin2008evaluating}
Keni Bernardin and Rainer Stiefelhagen.
\newblock Evaluating multiple object tracking performance: the clear mot
  metrics.
\newblock \emph{EURASIP Journal on Image and Video Processing}, 2008:\penalty0
  1--10, 2008.

\bibitem[Bewley et~al.(2016)Bewley, Ge, Ott, Ramos, and
  Upcroft]{bewley2016simple}
Alex Bewley, Zongyuan Ge, Lionel Ott, Fabio Ramos, and Ben Upcroft.
\newblock Simple online and realtime tracking.
\newblock In \emph{2016 IEEE international conference on image processing
  (ICIP)}, pages 3464--3468. IEEE, 2016.

\bibitem[Bochinski et~al.(2017)Bochinski, Eiselein, and
  Sikora]{bochinski2017high}
Erik Bochinski, Volker Eiselein, and Thomas Sikora.
\newblock High-speed tracking-by-detection without using image information.
\newblock In \emph{2017 14th IEEE international conference on advanced video
  and signal based surveillance (AVSS)}, pages 1--6. IEEE, 2017.

\bibitem[Botach et~al.(2022)Botach, Zheltonozhskii, and Baskin]{botach2022end}
Adam Botach, Evgenii Zheltonozhskii, and Chaim Baskin.
\newblock End-to-end referring video object segmentation with multimodal
  transformers.
\newblock In \emph{Proceedings of the IEEE/CVF Conference on Computer Vision
  and Pattern Recognition}, pages 4985--4995, 2022.

\bibitem[Cao et~al.(2023)Cao, Pang, Weng, Khirodkar, and
  Kitani]{cao2023observation}
Jinkun Cao, Jiangmiao Pang, Xinshuo Weng, Rawal Khirodkar, and Kris Kitani.
\newblock Observation-centric sort: Rethinking sort for robust multi-object
  tracking.
\newblock In \emph{Proceedings of the IEEE/CVF Conference on Computer Vision
  and Pattern Recognition}, pages 9686--9696, 2023.

\bibitem[Carion et~al.(2020)Carion, Massa, Synnaeve, Usunier, Kirillov, and
  Zagoruyko]{carion2020end}
Nicolas Carion, Francisco Massa, Gabriel Synnaeve, Nicolas Usunier, Alexander
  Kirillov, and Sergey Zagoruyko.
\newblock End-to-end object detection with transformers.
\newblock In \emph{European conference on computer vision}, pages 213--229.
  Springer, 2020.

\bibitem[Cho et~al.(2014)Cho, Van~Merri{\"e}nboer, Gulcehre, Bahdanau,
  Bougares, Schwenk, and Bengio]{cho2014learning}
Kyunghyun Cho, Bart Van~Merri{\"e}nboer, Caglar Gulcehre, Dzmitry Bahdanau,
  Fethi Bougares, Holger Schwenk, and Yoshua Bengio.
\newblock Learning phrase representations using rnn encoder-decoder for
  statistical machine translation.
\newblock \emph{arXiv preprint arXiv:1406.1078}, 2014.

\bibitem[Chun et~al.(2021)Chun, Oh, De~Rezende, Kalantidis, and
  Larlus]{chun2021probabilistic}
Sanghyuk Chun, Seong~Joon Oh, Rafael~Sampaio De~Rezende, Yannis Kalantidis, and
  Diane Larlus.
\newblock Probabilistic embeddings for cross-modal retrieval.
\newblock In \emph{Proceedings of the IEEE/CVF Conference on Computer Vision
  and Pattern Recognition}, pages 8415--8424, 2021.

\bibitem[Cui et~al.(2019)Cui, Jia, Lin, Song, and Belongie]{cui2019class}
Yin Cui, Menglin Jia, Tsung-Yi Lin, Yang Song, and Serge Belongie.
\newblock Class-balanced loss based on effective number of samples.
\newblock In \emph{Proceedings of the IEEE/CVF conference on computer vision
  and pattern recognition}, pages 9268--9277, 2019.

\bibitem[Du and Wu(2023)]{du2023no}
Yingxiao Du and Jianxin Wu.
\newblock No one left behind: Improving the worst categories in long-tailed
  learning.
\newblock In \emph{Proceedings of the IEEE/CVF Conference on Computer Vision
  and Pattern Recognition}, pages 15804--15813, 2023.

\bibitem[Du et~al.(2023{\natexlab{a}})Du, Shen, Zhen, and
  Snoek]{du2023superdisco}
Yingjun Du, Jiayi Shen, Xiantong Zhen, and Cees~GM Snoek.
\newblock Superdisco: Super-class discovery improves visual recognition for the
  long-tail.
\newblock In \emph{Proceedings of the IEEE/CVF Conference on Computer Vision
  and Pattern Recognition}, pages 19944--19954, 2023{\natexlab{a}}.

\bibitem[Du et~al.(2023{\natexlab{b}})Du, Zhao, Song, Zhao, Su, Gong, and
  Meng]{du2023strongsort}
Yunhao Du, Zhicheng Zhao, Yang Song, Yanyun Zhao, Fei Su, Tao Gong, and
  Hongying Meng.
\newblock Strongsort: Make deepsort great again.
\newblock \emph{IEEE Transactions on Multimedia}, 2023{\natexlab{b}}.

\bibitem[Geiger et~al.(2012)Geiger, Lenz, and Urtasun]{geiger2012we}
Andreas Geiger, Philip Lenz, and Raquel Urtasun.
\newblock Are we ready for autonomous driving? the kitti vision benchmark
  suite.
\newblock In \emph{2012 IEEE conference on computer vision and pattern
  recognition}, pages 3354--3361. IEEE, 2012.

\bibitem[Han et~al.(2022)Han, Huang, Wang, Yu, Liu, and Pan]{han2022mat}
Shoudong Han, Piao Huang, Hongwei Wang, En Yu, Donghaisheng Liu, and Xiaofeng
  Pan.
\newblock Mat: Motion-aware multi-object tracking.
\newblock \emph{Neurocomputing}, 476:\penalty0 75--86, 2022.

\bibitem[Hochreiter and Schmidhuber(1997)]{hochreiter1997long}
Sepp Hochreiter and J{\"u}rgen Schmidhuber.
\newblock Long short-term memory.
\newblock \emph{Neural computation}, 9\penalty0 (8):\penalty0 1735--1780, 1997.

\bibitem[Hyun~Cho and Kr{\"a}henb{\"u}hl(2022)]{hyun2022long}
Jang Hyun~Cho and Philipp Kr{\"a}henb{\"u}hl.
\newblock Long-tail detection with effective class-margins.
\newblock In \emph{European Conference on Computer Vision}, pages 698--714.
  Springer, 2022.

\bibitem[Jiang and Ye(2023)]{jiang2023cross}
Ding Jiang and Mang Ye.
\newblock Cross-modal implicit relation reasoning and aligning for
  text-to-image person retrieval.
\newblock In \emph{Proceedings of the IEEE/CVF Conference on Computer Vision
  and Pattern Recognition}, pages 2787--2797, 2023.

\bibitem[Jocher et~al.(2022)Jocher, Chaurasia, Stoken, Borovec, Kwon, Michael,
  Fang, Yifu, Wong, Montes, et~al.]{jocher2022ultralytics}
Glenn Jocher, Ayush Chaurasia, Alex Stoken, Jirka Borovec, Yonghye Kwon, Kalen
  Michael, Jiacong Fang, Zeng Yifu, Colin Wong, Diego Montes, et~al.
\newblock ultralytics/yolov5: v7. 0-yolov5 sota realtime instance segmentation.
\newblock \emph{Zenodo}, 2022.

\bibitem[Jocher et~al.(2023)Jocher, Chaurasia, and
  Qiu]{Jocher_YOLO_by_Ultralytics_2023}
Glenn Jocher, Ayush Chaurasia, and Jing Qiu.
\newblock {YOLO by Ultralytics}, 2023.

\bibitem[Kalman(1960)]{kalman1960new}
Rudolph~Emil Kalman.
\newblock A new approach to linear filtering and prediction problems.
\newblock 1960.

\bibitem[Li et~al.(2022)Li, Cao, and Zhang]{li2022learning}
Shiping Li, Min Cao, and Min Zhang.
\newblock Learning semantic-aligned feature representation for text-based
  person search.
\newblock In \emph{ICASSP 2022-2022 IEEE International Conference on Acoustics,
  Speech and Signal Processing (ICASSP)}, pages 2724--2728. IEEE, 2022.

\bibitem[Liang et~al.(2022{\natexlab{a}})Liang, Zhang, Zhou, Li, and
  Hu]{liang2022one}
Chao Liang, Zhipeng Zhang, Xue Zhou, Bing Li, and Weiming Hu.
\newblock One more check: making “fake background” be tracked again.
\newblock In \emph{Proceedings of the AAAI Conference on Artificial
  Intelligence}, pages 1546--1554, 2022{\natexlab{a}}.

\bibitem[Liang et~al.(2022{\natexlab{b}})Liang, Zhang, Zhou, Li, Zhu, and
  Hu]{liang2022rethinking}
Chao Liang, Zhipeng Zhang, Xue Zhou, Bing Li, Shuyuan Zhu, and Weiming Hu.
\newblock Rethinking the competition between detection and reid in multiobject
  tracking.
\newblock \emph{IEEE Transactions on Image Processing}, 31:\penalty0
  3182--3196, 2022{\natexlab{b}}.

\bibitem[Lin et~al.(2017)Lin, Goyal, Girshick, He, and
  Doll{\'a}r]{lin2017focal}
Tsung-Yi Lin, Priya Goyal, Ross Girshick, Kaiming He, and Piotr Doll{\'a}r.
\newblock Focal loss for dense object detection.
\newblock In \emph{Proceedings of the IEEE international conference on computer
  vision}, pages 2980--2988, 2017.

\bibitem[Luiten et~al.(2021)Luiten, Osep, Dendorfer, Torr, Geiger,
  Leal-Taix{\'e}, and Leibe]{luiten2021hota}
Jonathon Luiten, Aljosa Osep, Patrick Dendorfer, Philip Torr, Andreas Geiger,
  Laura Leal-Taix{\'e}, and Bastian Leibe.
\newblock Hota: A higher order metric for evaluating multi-object tracking.
\newblock \emph{International journal of computer vision}, 129:\penalty0
  548--578, 2021.

\bibitem[Ma et~al.(2022)Ma, Xu, Sun, Yan, Zhang, and Ji]{ma2022x}
Yiwei Ma, Guohai Xu, Xiaoshuai Sun, Ming Yan, Ji Zhang, and Rongrong Ji.
\newblock X-clip: End-to-end multi-grained contrastive learning for video-text
  retrieval.
\newblock In \emph{Proceedings of the 30th ACM International Conference on
  Multimedia}, pages 638--647, 2022.

\bibitem[Maggiolino et~al.(2023)Maggiolino, Ahmad, Cao, and
  Kitani]{maggiolino2023deep}
Gerard Maggiolino, Adnan Ahmad, Jinkun Cao, and Kris Kitani.
\newblock Deep oc-sort: Multi-pedestrian tracking by adaptive
  re-identification.
\newblock \emph{arXiv preprint arXiv:2302.11813}, 2023.

\bibitem[Meinhardt et~al.(2022)Meinhardt, Kirillov, Leal-Taixe, and
  Feichtenhofer]{meinhardt2022trackformer}
Tim Meinhardt, Alexander Kirillov, Laura Leal-Taixe, and Christoph
  Feichtenhofer.
\newblock Trackformer: Multi-object tracking with transformers.
\newblock In \emph{Proceedings of the IEEE/CVF conference on computer vision
  and pattern recognition}, pages 8844--8854, 2022.

\bibitem[Radford et~al.(2021)Radford, Kim, Hallacy, Ramesh, Goh, Agarwal,
  Sastry, Askell, Mishkin, Clark, et~al.]{radford2021learning}
Alec Radford, Jong~Wook Kim, Chris Hallacy, Aditya Ramesh, Gabriel Goh,
  Sandhini Agarwal, Girish Sastry, Amanda Askell, Pamela Mishkin, Jack Clark,
  et~al.
\newblock Learning transferable visual models from natural language
  supervision.
\newblock In \emph{International conference on machine learning}, pages
  8748--8763. PMLR, 2021.

\bibitem[Ristani et~al.(2016)Ristani, Solera, Zou, Cucchiara, and
  Tomasi]{ristani2016performance}
Ergys Ristani, Francesco Solera, Roger Zou, Rita Cucchiara, and Carlo Tomasi.
\newblock Performance measures and a data set for multi-target, multi-camera
  tracking.
\newblock In \emph{European conference on computer vision}, pages 17--35.
  Springer, 2016.

\bibitem[Sun et~al.(2020)Sun, Cao, Jiang, Zhang, Xie, Yuan, Wang, and
  Luo]{sun2020transtrack}
Peize Sun, Jinkun Cao, Yi Jiang, Rufeng Zhang, Enze Xie, Zehuan Yuan, Changhu
  Wang, and Ping Luo.
\newblock Transtrack: Multiple object tracking with transformer.
\newblock \emph{arXiv preprint arXiv:2012.15460}, 2020.

\bibitem[Sun et~al.(2022)Sun, Cao, Jiang, Yuan, Bai, Kitani, and
  Luo]{sun2022dancetrack}
Peize Sun, Jinkun Cao, Yi Jiang, Zehuan Yuan, Song Bai, Kris Kitani, and Ping
  Luo.
\newblock Dancetrack: Multi-object tracking in uniform appearance and diverse
  motion.
\newblock In \emph{Proceedings of the IEEE/CVF Conference on Computer Vision
  and Pattern Recognition}, pages 20993--21002, 2022.

\bibitem[Vaswani et~al.(2017)Vaswani, Shazeer, Parmar, Uszkoreit, Jones, Gomez,
  Kaiser, and Polosukhin]{vaswani2017attention}
Ashish Vaswani, Noam Shazeer, Niki Parmar, Jakob Uszkoreit, Llion Jones,
  Aidan~N Gomez, {\L}ukasz Kaiser, and Illia Polosukhin.
\newblock Attention is all you need.
\newblock \emph{Advances in neural information processing systems}, 30, 2017.

\bibitem[Wang et~al.(2020)Wang, Zheng, Liu, Li, and Wang]{wang2020towards}
Zhongdao Wang, Liang Zheng, Yixuan Liu, Yali Li, and Shengjin Wang.
\newblock Towards real-time multi-object tracking.
\newblock In \emph{European Conference on Computer Vision}, pages 107--122.
  Springer, 2020.

\bibitem[Wojke et~al.(2017)Wojke, Bewley, and Paulus]{wojke2017simple}
Nicolai Wojke, Alex Bewley, and Dietrich Paulus.
\newblock Simple online and realtime tracking with a deep association metric.
\newblock In \emph{2017 IEEE international conference on image processing
  (ICIP)}, pages 3645--3649. IEEE, 2017.

\bibitem[Wu et~al.(2023{\natexlab{a}})Wu, Han, Wang, Dong, Zhang, and
  Shen]{wu2023referring}
Dongming Wu, Wencheng Han, Tiancai Wang, Xingping Dong, Xiangyu Zhang, and
  Jianbing Shen.
\newblock Referring multi-object tracking.
\newblock In \emph{Proceedings of the IEEE/CVF Conference on Computer Vision
  and Pattern Recognition}, pages 14633--14642, 2023{\natexlab{a}}.

\bibitem[Wu et~al.(2023{\natexlab{b}})Wu, Wang, Zhang, Zhang, and
  Shen]{wu2023onlinerefer}
Dongming Wu, Tiancai Wang, Yuang Zhang, Xiangyu Zhang, and Jianbing Shen.
\newblock Onlinerefer: A simple online baseline for referring video object
  segmentation.
\newblock In \emph{Proceedings of the IEEE/CVF International Conference on
  Computer Vision}, pages 2761--2770, 2023{\natexlab{b}}.

\bibitem[Wu et~al.(2022)Wu, Jiang, Sun, Yuan, and Luo]{wu2022language}
Jiannan Wu, Yi Jiang, Peize Sun, Zehuan Yuan, and Ping Luo.
\newblock Language as queries for referring video object segmentation.
\newblock In \emph{Proceedings of the IEEE/CVF Conference on Computer Vision
  and Pattern Recognition}, pages 4974--4984, 2022.

\bibitem[Yan et~al.(2022)Yan, Dong, Zhang, and Tang]{yan2022clip}
Shuanglin Yan, Neng Dong, Liyan Zhang, and Jinhui Tang.
\newblock Clip-driven fine-grained text-image person re-identification.
\newblock \emph{arXiv preprint arXiv:2210.10276}, 2022.

\bibitem[Yan et~al.(2023)Yan, Tang, Zhang, and Tang]{yan2023image}
Shuanglin Yan, Hao Tang, Liyan Zhang, and Jinhui Tang.
\newblock Image-specific information suppression and implicit local alignment
  for text-based person search.
\newblock \emph{IEEE Transactions on Neural Networks and Learning Systems},
  2023.

\bibitem[Yang et~al.(2023)Yang, Odashima, Masui, and Jiang]{yang2023hard}
Fan Yang, Shigeyuki Odashima, Shoichi Masui, and Shan Jiang.
\newblock Hard to track objects with irregular motions and similar appearances?
  make it easier by buffering the matching space.
\newblock In \emph{Proceedings of the IEEE/CVF Winter Conference on
  Applications of Computer Vision}, pages 4799--4808, 2023.

\bibitem[Yu et~al.(2016)Yu, Li, Li, Liu, Shi, and Yan]{yu2016poi}
Fengwei Yu, Wenbo Li, Quanquan Li, Yu Liu, Xiaohua Shi, and Junjie Yan.
\newblock Poi: Multiple object tracking with high performance detection and
  appearance feature.
\newblock In \emph{Computer Vision--ECCV 2016 Workshops: Amsterdam, The
  Netherlands, October 8-10 and 15-16, 2016, Proceedings, Part II 14}, pages
  36--42. Springer, 2016.

\bibitem[Yu et~al.(2018)Yu, Wang, Shelhamer, and Darrell]{yu2018deep}
Fisher Yu, Dequan Wang, Evan Shelhamer, and Trevor Darrell.
\newblock Deep layer aggregation.
\newblock In \emph{Proceedings of the IEEE conference on computer vision and
  pattern recognition}, pages 2403--2412, 2018.

\bibitem[Zhang et~al.(2023{\natexlab{a}})Zhang, Wang, Zhang, Zhang, and
  Zhong]{zhang2023one}
Huanlong Zhang, Jingchao Wang, Jianwei Zhang, Tianzhu Zhang, and Bineng Zhong.
\newblock One-stream vision-language memory network for object tracking.
\newblock \emph{IEEE Transactions on Multimedia}, 2023{\natexlab{a}}.

\bibitem[Zhang et~al.(2021)Zhang, Wang, Wang, Zeng, and Liu]{zhang2021fairmot}
Yifu Zhang, Chunyu Wang, Xinggang Wang, Wenjun Zeng, and Wenyu Liu.
\newblock Fairmot: On the fairness of detection and re-identification in
  multiple object tracking.
\newblock \emph{International Journal of Computer Vision}, 129:\penalty0
  3069--3087, 2021.

\bibitem[Zhang et~al.(2022)Zhang, Sun, Jiang, Yu, Weng, Yuan, Luo, Liu, and
  Wang]{zhang2022bytetrack}
Yifu Zhang, Peize Sun, Yi Jiang, Dongdong Yu, Fucheng Weng, Zehuan Yuan, Ping
  Luo, Wenyu Liu, and Xinggang Wang.
\newblock Bytetrack: Multi-object tracking by associating every detection box.
\newblock In \emph{European Conference on Computer Vision}, pages 1--21.
  Springer, 2022.

\bibitem[Zhang et~al.(2023{\natexlab{b}})Zhang, Wang, and
  Zhang]{zhang2023motrv2}
Yuang Zhang, Tiancai Wang, and Xiangyu Zhang.
\newblock Motrv2: Bootstrapping end-to-end multi-object tracking by pretrained
  object detectors.
\newblock In \emph{Proceedings of the IEEE/CVF Conference on Computer Vision
  and Pattern Recognition}, pages 22056--22065, 2023{\natexlab{b}}.

\bibitem[Zhao et~al.(2023{\natexlab{a}})Zhao, Wang, Wang, Lu, and
  Ruan]{zhao2023transformer}
Haojie Zhao, Xiao Wang, Dong Wang, Huchuan Lu, and Xiang Ruan.
\newblock Transformer vision-language tracking via proxy token guided
  cross-modal fusion.
\newblock \emph{Pattern Recognition Letters}, 168:\penalty0 10--16,
  2023{\natexlab{a}}.

\bibitem[Zhao et~al.(2023{\natexlab{b}})Zhao, Jiang, Hu, Zhang, and
  Liu]{zhao2023mdcs}
Qihao Zhao, Chen Jiang, Wei Hu, Fan Zhang, and Jun Liu.
\newblock Mdcs: More diverse experts with consistency self-distillation for
  long-tailed recognition.
\newblock In \emph{Proceedings of the IEEE/CVF International Conference on
  Computer Vision}, pages 11597--11608, 2023{\natexlab{b}}.

\bibitem[Zheng et~al.(2023)Zheng, Zhong, Liang, Li, Ji, and
  Li]{zheng2023towards}
Yaozong Zheng, Bineng Zhong, Qihua Liang, Guorong Li, Rongrong Ji, and Xianxian
  Li.
\newblock Towards unified token learning for vision-language tracking.
\newblock \emph{IEEE Transactions on Circuits and Systems for Video
  Technology}, 2023.

\bibitem[Zhou et~al.(2023)Zhou, Zhou, Mao, and He]{zhou2023joint}
Li Zhou, Zikun Zhou, Kaige Mao, and Zhenyu He.
\newblock Joint visual grounding and tracking with natural language
  specification.
\newblock In \emph{Proceedings of the IEEE/CVF Conference on Computer Vision
  and Pattern Recognition}, pages 23151--23160, 2023.

\bibitem[Zhu et~al.(2020)Zhu, Su, Lu, Li, Wang, and Dai]{zhu2020deformable}
Xizhou Zhu, Weijie Su, Lewei Lu, Bin Li, Xiaogang Wang, and Jifeng Dai.
\newblock Deformable detr: Deformable transformers for end-to-end object
  detection.
\newblock \emph{arXiv preprint arXiv:2010.04159}, 2020.

\end{thebibliography}


\begin{thebibliography}{0}
\providecommand{\natexlab}[1]{#1}
\providecommand{\url}[1]{\texttt{#1}}
\expandafter\ifx\csname urlstyle\endcsname\relax
  \providecommand{\doi}[1]{doi: #1}\else
  \providecommand{\doi}{doi: \begingroup \urlstyle{rm}\Url}\fi

\end{thebibliography}
